\documentclass[10pt,twocolumn,letterpaper]{article}

\usepackage{wacv}
\usepackage{times}
\usepackage{epsfig}
\usepackage{graphicx}
\usepackage{amsmath}
\usepackage{amssymb}
\usepackage{booktabs}
\usepackage{xcolor}
\usepackage{enumitem}
\usepackage{lipsum}
\usepackage[ruled,vlined,linesnumbered]{algorithm2e}
\usepackage{physics}
\usepackage{mathtools}
\usepackage{appendix}

\definecolor{codegreen}{rgb}{0.0,0.6,0.0}

\SetCommentSty{mycommfont}
\usepackage{float} 
\usepackage{caption}
\usepackage{datetime}

\makeatletter
\newcommand{\algorithmfootnote}[2][\footnotesize]{%
  \let\old@algocf@finish\@algocf@finish
  \def\@algocf@finish{\old@algocf@finish
    \leavevmode\rlap{\begin{minipage}{\linewidth}
    #1#2
    \end{minipage}}%
  }%
}
\makeatother

\usepackage[pagebackref=true,breaklinks=true,letterpaper=true,colorlinks,bookmarks=false]{hyperref}

\wacvfinalcopy 

\def\wacvPaperID{****} 

\title{Deep HM-SORT: Enhancing Multi-Object Tracking in Sports with Deep Features, Harmonic Mean, and Expansion IOU}

\begin{document}

\author{
Matias Gran-Henriksen\thanks{Contributed equally to this paper.}, Hans Andreas Lindgård\footnotemark[1], Gabriel Kiss, Frank Lindseth \\
Norwegian University of Science and Technology - NTNU \\
{\tt\small \{matiasgr, hansal\}@stud.ntnu.no}
}
\twocolumn[{%
\renewcommand\twocolumn[1][]{#1}%
\footnotetext{These authors contributed equally to this work.}

\maketitle
\ifwacvfinal\thispagestyle{empty}\fi
}]
\footnotetext{These authors contributed equally to this work.}

\begin{abstract}
This paper introduces Deep HM-SORT, a novel online multi-object tracking algorithm specifically designed to enhance the tracking of athletes in sports scenarios. Traditional multi-object tracking methods often struggle with sports environments due to the similar appearances of players, irregular and unpredictable movements, and significant camera motion. Deep HM-SORT addresses these challenges by integrating deep features, harmonic mean, and Expansion IOU. By leveraging the harmonic mean, our method effectively balances appearance and motion cues, significantly reducing ID-swaps. Additionally, our approach retains all tracklets indefinitely, improving the re-identification of players who leave and re-enter the frame.

Experimental results demonstrate that Deep HM-SORT achieves state-of-the-art performance on two large-scale public benchmarks, SportsMOT and SoccerNet Tracking Challenge 2023. Specifically, our method achieves 80.1 HOTA on the SportsMOT dataset and 85.4 HOTA on the SoccerNet-Tracking dataset, outperforming existing trackers in key metrics such as HOTA, IDF1, AssA, and MOTA. This robust solution provides enhanced accuracy and reliability for automated sports analytics, offering significant improvements over previous methods without introducing additional computational cost.

\end{abstract}

\section{Introduction}
The aim of Multiple Object Tracking (MOT) is to identify and monitor all relevant objects within a video sequence. MOT is a crucial area of study with diverse applications, including sports analytics and wildlife monitoring. Recent tracking algorithms\cite{botSORT, byteTrack, strongSORT} predominantly concentrate on tracking pedestrians or vehicles and have demonstrated strong performance on public benchmarks like MOT16\cite{mot16}. However, despite their success in these areas, these trackers struggle with more challenging benchmarks, particularly those related to sports scenarios\cite{sportsMOT}. The increasing demand for automated sports analytics, including tactical analysis and player movement statistics, highlights the need for greater focus on applying MOT to sports.

The way players move in sports makes it extremely challenging to maintain robust tracking over long periods. This difficulty arises mainly due to players' irregular movements and occlusions, but also because of other factors such as significant camera movements. Various methods have been attempted to handle these irregular movements by moving away from traditional IOU, which measures the overlap between a track's predicted position and the location of a detection. Numerous studies \cite{ocSORT, deepEIoU, CBIOU} have highlighted IOU's shortcomings in effectively handling non-linear and rapid motion. Standard IOU metrics yield subpar results in football tracking, where players exhibit irregular and large movements compounded by potential camera motion.

However, what is not as widely addressed in the challenge of sports tracking is the re-identification process, where players often look very similar. Players on each team wear identical socks, shorts, and jerseys, making it difficult to distinguish them. Additionally, the footage is typically zoomed out, complicating the task of capturing fine details on discriminating features, such as players' faces and jersey numbers. When players move closely together, it can also confuse the tracker and result in ID swaps. This leads to the biggest challenge for fully robust tracking within sports: the ability to hold consistent tracks on players throughout a sequence.

To address these challenges, we propose a novel and robust online multi-object tracking algorithm designed for holding consistent tracking of players within sports and reduce ID-swaps. Our experimental results show that our algorithm drastically reduces the ID-swaps. It also outperforms all existing tracking algorithms on  SportsMOT test set \cite{sportsMOT}, while remaining an online tracking method. This paper makes three main contributions:

\begin{itemize}
    \item We present a novel association method to address the specific challenge of ID-swaps in sports tracking, named Deep HM-SORT. This simple online tracking algorithm is designed specifically for sports scenarios. The proposed method achieves an 80.1 HOTA on the SportsMOT dataset \cite{sportsMOT}, significantly reducing the number of ID-swaps. It showcases state-of-the-art performance by outperforming all previously established tracking algorithms on the SportsMOT dataset \cite{sportsMOT}.
    \item Our new method of combining the costs from feature embeddings and IOU with harmonic mean to reduce the number of ID-swaps.
    \item Address that sports occur in a closed environment and demonstrate improved tracking performance by retaining all tracks within a tracking sequence. This approach enhances tracking accuracy and enables the re-identification of players who have been out of the frame for extended periods.
\end{itemize}

\section{Related Work} 
\subsection{Multiple Object Tracking using Kalman Filter}
The majority of existing tracking algorithms \cite{SORT, ocSORT, botSORT, deepSORT, strongSORT} employ a Kalman Filter (KF) \cite{kalmanFilter} to model object motion. The KF predicts an object's position in the next frame based on its motion in previous frames. While many KF-based trackers have demonstrated strong performance on public benchmarks \cite{SORT, ocSORT}, the KF's linear motion and Gaussian noise assumptions can lead to inaccuracies in predicting an object's location during large or non-linear movements. To address this issue, several trackers have modified the KF. For instance, BoT-SORT increases the size of the KF vector to include estimating both width and height directly, while OC-SORT adjusts the KF parameters during tracking to prevent error accumulation \cite{ocSORT, deepocsort, botSORT}.

\subsubsection{Intersection Over Union}
\begin{figure}[htbp]
    \centering
    \includegraphics[width=0.75\linewidth]{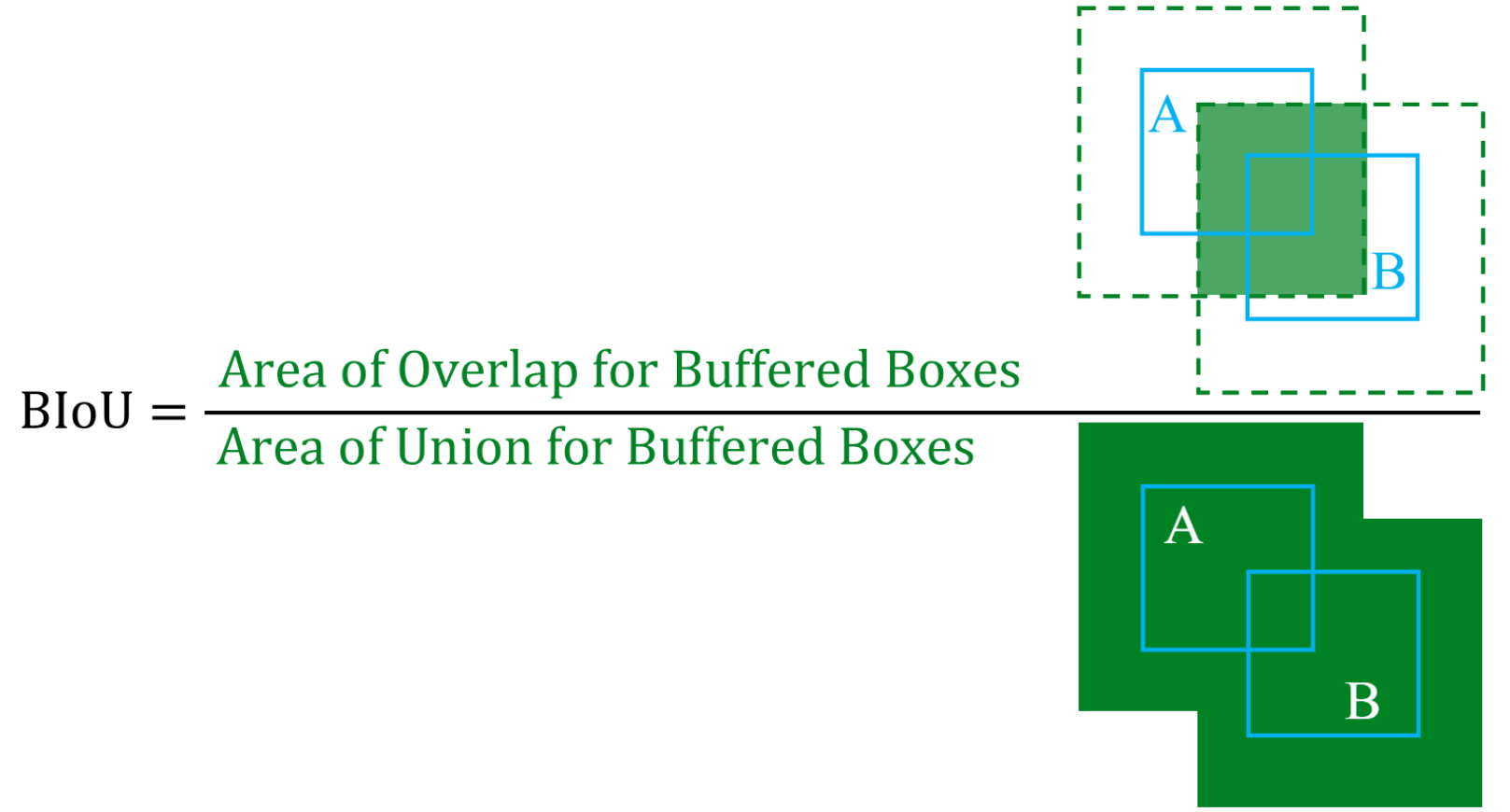}
    \caption[Expanded IoU]{Showing the IoU between two expanded bounding boxes. Source \cite{CBIOU}.}
    \label{fig:bufferedIoU}
\end{figure}

To associate tracks with detections in a KF Intersection over Union (IoU) is commonly used. This value is determined between the predicted location of a track and the position of a detected object. After constructing a complete cost matrix of size $m$ tracks by $n$ detections ($m\times n$), the Hungarian Algorithm is employed to find the most optimal assignment betweem each track and detection. Recent work within sports tracking \cite{CBIOU,deepEIoU} has, however, utilized buffered or expanded IOU, demonstrated in Figure \ref{fig:bufferedIoU}, to compensate for the irregular movements found in sports.

\subsection{Appearance-based Multiple Object Tracking}
Advancements in deep learning have enabled the use of the Re-Id module, traditionally used for re-identifying lost objects, in the association stage of object tracking\cite{deepEIoU,deepocsort,deepSORT}. This process typically involves calculating the cosine distance between a track's feature vector and a detection's feature vector and using this distance for association. By leveraging the appearance features of an object, the tracking process becomes more robust. Modern tracking approaches often combine data from both KF-based tracks and appearance-based tracks to more accurately assign IDs .

Appearance-based tracking methods enhance tracking robustness with additional appearance clues. However, these methods can sometimes be unreliable due to factors such as occlusions, similar appearances among tracked objects, or varying lighting conditions. BoT-SORT, for instance, calculates the cosine distance in the appearance space between a detection and a track, then computes the IoU distance, and uses the minimum of these two values for association, leading to impressive results at its release.

\subsection{Multi-Object Tracking in Sports}
Deep EIoU (June 2023) expands upon the methodology of BoT-SORT \cite{botSORT} to achieve the SOTA score on SportsMOT. Huang et al. \cite{deepEIoU} present that other SOTA algorithms perform poorly on difficult benchmarks, especially those containing sports scenarios, partly due to the fast and unpredictable motion of tracked objects in sports. Traditional IOU is a fundamental component of location-based tracking methods. However, it is not flexible enough to accommodate large object movements. This limitation can result in the track and bounding box of the same object having no IOU in consecutive frames. Central to the algorithm presented is the iterative expansion of bounding boxes. Bounding boxes that initially don't find a matching track have their bounding boxes expanded by a scaling factor $e$. The Hungarian algorithm is then used for association using the expanded bounding boxes. The above process is repeated twice with the expansion scale increased from $e$ to $e+a$, with $a<e$.

Another big contribution within tracking in sports is Maglo et al. \cite{SnTrackingWinner}, which addresses ID-swapping during player occlusions by splitting the association process into two distinct stages. Unlike previous methods that combine IoU and embedding distances, their approach relies heavily on a re-ID module.

Initially, the method uses a YOLOX detection model to assign IDs to detections. As long as players are not occluded, tracking is based on IoU distance and a distance criterion calculated from players' relative positions on the field, using a homography model to account for camera movement. This matching is performed with SORT.

When occlusions occur, active tracks are halted, and new ones are started. The re-identification stage then begins, using a pre-trained re-identification network, further fine-tuned on player thumbnails from previous tracklets. Iterative associations are made using cost matrices based on mean Euclidean distances and the Hungarian Algorithm, reducing ID-swapping but resulting in slower, offline tracking.
\section{Proposed Method}
In this section, we present our two main modifications and improvements for multi-object tracking-based tracking-by-detection and, more specifically, tracking within sports. By integrating these changes into DeepEIOU, we present a new tracker with state-of-the-art performance in SportsMOT.

\subsection{IoU - Re-ID Fusion}
!Skirve noe om hvordan feature vector blir oppdatert!
The feature vector used for association is updated whenever a new detection is matched to that track. It is updated according to Equation \ref{appearance_update}. Where the feature vector for track $i$ in frame $k$ $e_i^k$ 

\begin{equation}
\label{appearance_update}
    v_i^k = \alpha v_i^{k-1} + (1-\alpha) f_i^k
\end{equation}

The tracker implementation draws heavily on the framework developed by Huang et al.\cite{deepEIoU}, specifically their method of iterative IoU scaling. However, upon review, we identified opportunities for optimization in their approach to constructing the association cost matrix. Deep-EIoU, which extends BoT-SORT, selects values for the cost matrix by taking the minimum of the EIoU distance and the embedding distance. While this method has shown good results, it can be less effective in scenarios where both costs are low.

We replace this approach, and instead utilize the harmonic mean to calculate a more balanced cost in the cost matrix. The harmonic mean, shown in Equation \ref{eq:harmonic_mean}, emphasizes the smaller of two values, thereby providing an average that skews towards the lower of the two measurements. Therefore, for each pair of track $j$ and detection $k$, we compute the harmonic mean of their EIoU distance $d_{j,k}^1$ (\ref{eq:eiou}) and embedding distance (cosine distance, \ref{eq:cosine_dist}) $d_{j,k}^2$ to create the cost matrix.

\begin{equation}
     d_{j,k}^1 = 1 - EIoU(j, k)
    \label{eq:eiou}
\end{equation}

\begin{equation}
    d_{j,k}^2 = 1 - \frac{ j \cdot k}{| j| | k|}
    \label{eq:cosine_dist}
\end{equation}

\begin{equation}
     H(j, k) = \frac{2}{\dfrac{1}{d_{j,k}^1}+\dfrac{1}{d_{j,k}^2}}
    \label{eq:harmonic_mean}
\end{equation}

As with other methods\cite{botSORT, deepEIoU} that utilize the lowest cost from either appearance features or IOU, relying on only one of these criteria can result in a weaker foundation for making accurate object associations between frames, particularly when both methods provide strong cues. This is especially important in sports tracking, where players on the same team wear identical equipment and uniforms. In situations where two players from the same team come close to each other, their embedding vectors can become quite similar. When associating new detections, ID-swaps can easily occur if the association is based solely on appearance criteria. The advantage of using the harmonic mean is that it considers both appearance and motion when associating tracks. This approach can reduce ID-swaps in scenarios where players look alike, as considering IOU can provide valuable guidance for making the correct association.

\begin{figure}[h]
    \centering
    \includegraphics[width=0.8\linewidth]{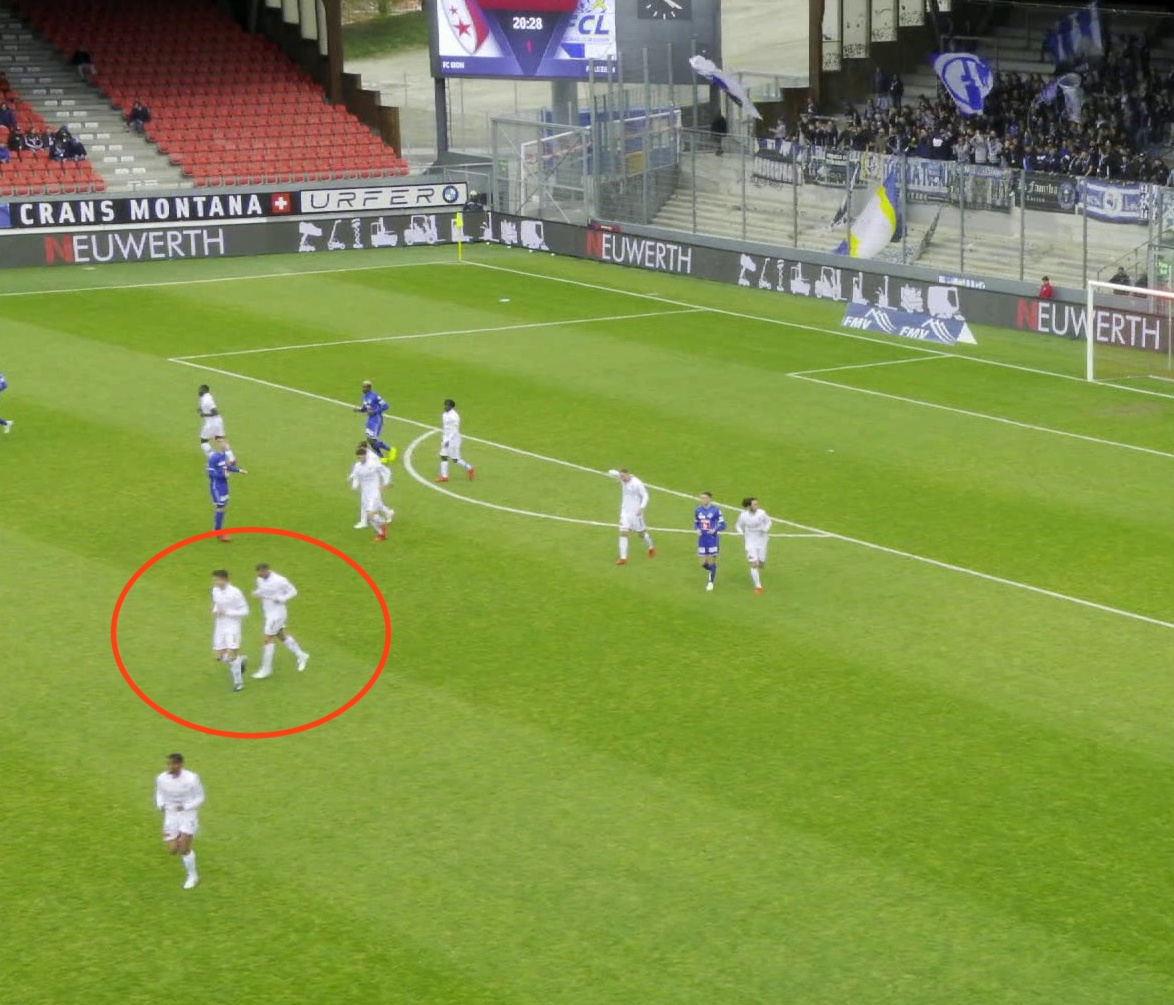}
    \caption{Demonstrating two similar players from the SoccerNet dataset \cite{soccerNetPaper} before they are about to cross each other, potentially leading to an ID-swap.}
    \label{fig:player_sim}
\end{figure}

\subsection{Tracklet Management}
Another area for improvement that has not been widely addressed is sports tracking, which occurs in a closed environment. Therefore, Deep HM-SORT has moved away from using a caped temporary ID buffer, which many trackers utilize \cite{botSORT, byteTrack, deepEIoU, deepSORT}. As setting a maximum number of IDs hindered optimal association, we have taken an alternative approach: not discarding ID tracklets from the buffer. Instead, we preserve all tracklets indefinitely. This approach allows the tracker to more easily associate players who disappear and then return to the camera's view. Players who leave and re-enter the frame often return to a similar position within the frame, which can now be easier associated with a bigger buffered IOU together with the re-ID module.

\section{Experiments}
\begin{table*}[t]
\centering
\begin{tabular}{lcccccccccc}
\toprule
\textbf{Method} & \textbf{Training Setup} & \textbf{HOTA↑} & \textbf{IDF1↑} & \textbf{AssA↑} & \textbf{MOTA↑} & \textbf{DetA↑} & \textbf{LocA↑} & \textbf{IDs↓} & \textbf{Frag↓} \\
\midrule
FairMOT \cite{fairmot} & Train & 49.3 & 53.5 & 34.7 & 86.4 & 70.2 & 83.9 & 9928 & 21673 \\
QDTrack \cite{qdtrack} & Train & 60.4 & 62.3 & 47.2 & 90.1 & 77.5 & 88.0 & 6377 & 11850 \\
CenterTrack \cite{centertrack} & Train & 62.7 & 60.0 & 48.0 & 90.8 & 82.1 & 90.8 & 10481 & 5750 \\
TransTrack \cite{transtrack} & Train & 68.9 & 71.5 & 57.5 & 92.6 & 82.7 & 91.0 & 4992 & 9994 \\
BoT-SORT \cite{botSORT}& Train & 68.7 & 70.0 & 55.9 & 94.5 & 84.4 & 90.5 & 6729 & 5349 \\
ByteTrack \cite{byteTrack} & Train & 62.8 & 69.8 & 51.4 & 94.5 & 77.1 & 85.6 & 3267 & 4499 \\
OC-SORT \cite{ocSORT} & Train & 71.9 & 72.2 & 59.4 & 95.4 & 86.4 & 92.4 & 3093 & 3474 \\
ByteTrack \cite{byteTrack} & Train+Val & 64.1 & 71.4 & 52.3 & 95.9 & 78.5 & 85.7 & 3089 & 4216 \\
OC-SORT \cite{ocSORT} & Train+Val & 73.7 & 74.0 & 61.5 & 96.5 & \textbf{88.5} & 92.7 & 2728 & 3144 \\
MixSort-Byte \cite{sportsMOT} & Train+Val & 65.7 & 74.1 & 54.8 & 96.2 & 78.8 & 85.7 & 2472 & 4009 \\
MixSort-OC \cite{sportsMOT} & Train+Val& 74.1 & 74.4 & 62.0 & 96.5 & \textbf{88.5} & \textbf{92.7} & 2781 & 3199 \\
MeMOTR (D-DETR) \cite{memotr} & Train & 68.8 & - & 57.8 & - & 82.0 & - & - & - \\
MeMOTR \cite{memotr} & Train & 70.0 & - & 59.1 & - & 83.1 & - & - & - \\
MotionTrack \cite{motiontrack} & Train & 74.0 & - & 61.7 & - & 74.0 & - & - & - \\
MoveSORT \cite{movesort} & Train+Val & 74.6 & - & 63.7 & - & 87.5 & - & - & - \\
Deep-EIoU \cite{deepEIoU} & Train & 74.1 & 75.0 & 63.1 & 95.1 & 87.2 & 92.5 & 3066 & 3471 \\
Deep-EIoU \cite{deepEIoU} & Train+Val & 77.2 & 79.8 & 67.7 & 96.3 & 88.2 & 92.4 & 2659 & 3081 \\
\midrule
Deep HM-SORT (Ours) & Train+Val & \textbf{80.1} & \textbf{85.2} & \textbf{72.7} & \textbf{96.6} & 88.3 & 92.4 & \textbf{1703} & \textbf{2996} \\
\bottomrule
\end{tabular}
\caption{Comparison of different trackers on the SportsMOT test set. The evaluation results from MoveSORT, MotionTrack, and MeMOTR are taken from Papers With Code. The rest are reported from the results of Huang et al.\cite{deepEIoU}. Deep HM-SORT outperforms all previous reported trackers on HOTA, IDF1, AssA, MOTA, ID swaps, and Fragmentation, making it SOTA on the most significant metrics for MOT on SportsMOT}
\label{tab:comparison}
\end{table*}

\subsection{Experimental Settings}

\noindent \textbf{Datasets.}
Testing was performed on two popular benchmarks within sports tracking: SportsMOT, and SoccerNet Tracking Challenge 2023. SportsMOT contains sequences filmed from a single moving camera across three sports, football, basketball, and volleyball, which consists of a total of 240 video clips from the three categories, the test set contains 150 clips, varying between 250 and 3600 frames each. SoccerNet Tracking Challenge 2023 contains similar sequences, except only for football. \\

\noindent \textbf{Metrics.} 
Evaluations were performed according to the standard CLEAR and Higher Order Tracking Accuracy (HOTA) metrics. Including Multiple Object Tracking Accuracy (MOTA), ID Switch (IDSW), Fragmented Tracks (Frag), etc. HOTA is the metric mainly in focus, as it balances detection, association, and localization accuracy into a single unified metric. We also place a strong emphasis on preventing ID swaps. The primary purpose of this tracker is to facilitate accurate sport statistics at the individual level. For these statistics to be meaningful, it's crucial that ID swaps do not occur, as they would result in incorrect data for individual players. \\

\noindent \textbf{Implementation details.}
All the experiments were implemented using PyTorch and ran on a Virtual Machine (VM), which was gained access to through Norwegian University of Science and Technology (NTNU). This VM was outfitted with a NVIDIA L40-16Q GPU with CUDA version 12.0. For SportsMOT we employ the YOLOX model trained by \cite{deepEIoU}. Whilst for SoccerNet we use a YOLOv8-XL trained for 150 epochs on our own dataset. Finally for both these two benchmarks we use the feature extractor model trained by \cite{deepEIoU}. The tracking hyperparameters are the same for all of the evaluations. The detection score threshold was set to 0.4. We define high detection score threshold used for the initial round of association to be 0.6, while the threshold below which we discard tracks is set to 0.5. In linear assignment if the detection and track assignment cost is higher than 0.8, the cost is set to 1. Similarly if the appearance based assignment cost between a detection and track is higher than 0.3, the cost is set to 1. When expanding bounding boxes we start with an initial expansion of 0.3, and increase it by 0.3, after the initial association step.

\subsection{Ablation Study}
To verify the performance of both Harmonic Mean for association and not removing tracks, we conduct an ablation study on the SportsMOT test set and SoccerNet Tracking Challenge 2023 by integrating these two approaches into Deep EIoU. We maintain consistent hyperparameter settings across all four tests. Initially, we test with only Deep EIoU using the hyperparameters. Next, we test by keeping all tracks without Harmonic Mean, then with Harmonic Mean and track removal, and finally by applying both simultaneously. For SportsMOT the results clearly show the impact of these improvements. Keeping all tracks significantly increases the HOTA score, while implementing Harmonic Mean notably reduces the number of ID switches. Using both methods together increases the HOTA score further and reduces ID switches even more. On SoccerNet we don't have access to ID swap metrics, but here harmonic mean and not removing tracks both increase the HOTA, DetA, and AssA score individually, and the best result is achieved by applying both. This ablation study shows the efficacy of both approaches presented in this paper.

\begin{table}[H]
\centering
  \begin{tabular}{ccccc}
    \toprule
    \textbf{HM} & \textbf{Keep all tracks} & \textbf{HOTA↑} & \textbf{IDSW↓} \\
    \midrule
                 &               & 77.5          & 2721 \\
                 & $\checkmark$  & \textbf{80.2} & 2418 \\
    $\checkmark$ &               & 77.21         & 2146 \\
    $\checkmark$ & $\checkmark$  & 80.1 & \textbf{1703} \\
    \bottomrule
  \end{tabular}
  \caption{Evaluation of Deep-EIoU with and without our improvements on the SportsMOT test set. Experiment results show that incorporating not removing tracks benefits the most, while harmonic mean is not positive on the HOTA score, but massively improves ID-swaps.}
  \label{tab:your_table_label}
\end{table}

\begin{table}[H]
\centering
  \begin{tabular}{cccccc}
    \toprule
    \textbf{HM} & \textbf{Keep all tracks} & \textbf{HOTA↑} & \textbf{DetA↑} & \textbf{AssA↑} \\
    \midrule
                 &               & 67.34  & 70.83  & 64.11 \\
                 & $\checkmark$  & 68.10  & 71.11  & 65.30 \\
    $\checkmark$ &               & 68.87  & 71.13  & 66.76 \\
    $\checkmark$ & $\checkmark$  & \textbf{69.04}  & \textbf{71.35}  & \textbf{66.87} \\
    \bottomrule
  \end{tabular}
  \caption{Evaluation of Deep-EIoU with and without our improvements on SoccerNet Tracking Challenge 2023. The experimental results show that both innovations improve the HOTA score, and that harmonic mean improves the HOTA the most.}
  \label{tab:your_table_label}
\end{table}

\subsection{Benchmarks Evaluation}
We compare Deep HM-SORT to other trackers on SportsMOT in Table \ref{tab:comparison}, SoccerNet Tracking Challenge 2023 in Table \ref{tab:soccernet2023ours}. All the results are obtained from either SoccerNet for SoccerNet Tracking Challenge 2023, or \href{paperswithcode.com}{paperswithcode.com} for SportsMOT. \\

\noindent \textbf{SportsMOT.} 
Deep HM-SORT outperforms all other SOTA trackers in all the main metrics, i.e. HOTA and MOTA. The only metrics we don't get the SOTA performance is in DetA and LocA, which is comparable to the others scores, only being 0.2 and 0.3 behind respectively. In the other metrics we improve massively upon the previous scores, achieving a 2.9\% higher HOTA and over 900 fewer ID swaps, which is a 30\% reduction over the previous best. \\

\noindent \textbf{SoccerNet.} 
We achieved 3rd place in the SoccerNet Tracking Challenge 2023. The benchmark results for this challenge are less detailed compared to SportsMOT, preventing us from evaluating our performance on metrics like IDSW. However, we rank 2nd in AssA, surpassing the overall 2nd place team in this metric. With a better detection model specifically trained for this challenge, we likely would have secured 2nd place overall. The 1st place team uses a slow offline method, taking over 16 minutes to process a 30-second clip. In contrast, our approach is online and much faster, taking only 1 minute and 30 seconds for the same clip. This demonstrates the inherent trade-off between speed and accuracy.

\begin{table}[h]
    \centering
    \begin{tabular}{lccc}
    \toprule
    \textbf{Team} & \textbf{HOTA↑} & \textbf{DetA↑} & \textbf{AssA↑} \\
    \midrule
    Baseline & 42.38 & 34.41 & 52.21 \\
    scnu & 58.07 & 64.77 & 52.23 \\
    ZTrackers & 58.69 & 68.69 & 50.25 \\
    SAIVA\_Tracking & 63.2 & 70.45 & 56.87 \\
    ICOST & 65.67 & 73.07 & 59.17 \\
    MOT4MOT & 66.27 & 70.32 & 62.62 \\ \midrule
    Ours & 69.04 & 71.35 & 66.87 \\ \midrule
    MTIOT & 69.54 & 75.18 & 64.45 \\ 
    Kalisteo & \textbf{75.61} & \textbf{75.38} & \textbf{75.94} \\
    \bottomrule
    \end{tabular}
    \caption[SoccerNet Leaderboard]{Scores from the 2023 SoccerNet tracking challenge (Detection and Tracking). Source \cite{SoccerNet2023}.}
    \label{tab:soccernet2023ours}
\end{table}

\subsection{Limitations}
Deep HM-SORT still has some limitations. In occlusion events the appearance of a player is affected by the player they are occluded by or occluding, this affects the appearance based matching as the appearance vector is affected by the everything in the thumbnail. This often leads to a temporary ID-swap, however most times the players swap back when they are not occluding each other anymore. Another limitation is the run time. When running both object detection inference and Re-Id inference the FPS is too low to be used in real-time scenarios, on our hardware capping out around 13 FPS. Lastly, as this tracker is an online approach there are no offline post-processing trajectory refinement methods. Implementing offline approaches for handling occlusion such as \cite{SnTrackingWinner} would likely further improve the tracker.

\section{Conclusion}
In this paper we propose an enhanced multiple object tracking approach using Harmonic Mean and a semi-closed environment, named Deep HM-SORT, which is SOTA on SportsMOT, ranking 1st on HOTA, IDF1, AssA, MOTA, IDsw, and frag. We are also able to achieve a 3rd place on the SoccerNet Tracking Challenge 2023, just behind 2nd place. The proposed method can be easily integrated into other tracking-by-detection trackers to increase robustness against ID swaps.

\section{Acknowledgement}
We would like to thank the Rosenborg Ballklubb team which has provided us raw data and help in understanding their football tracking needs. We also appreciate the Norwegian University of Science and Technology for providing the necessary resources and facilities for conducting this research.

{\small
\bibliographystyle{ieee}
\bibliography{egbib}

\begin{thebibliography}{10}\itemsep=-1pt

\bibitem{movesort}
M.~Adžemović, P.~Tadić, A.~Petrović, and M.~Nikolić.
\newblock Beyond kalman filters: Deep learning-based filters for improved object tracking, 2024.

\bibitem{botSORT}
N.~Aharon, R.~Orfaig, and B.-Z. Bobrovsky.
\newblock Bot-sort: Robust associations multi-pedestrian tracking.
\newblock {\em arXiv preprint arXiv:2206.14651}, 2022.

\bibitem{SORT}
A.~Bewley, Z.~Ge, L.~Ott, F.~Ramos, and B.~Upcroft.
\newblock Simple online and realtime tracking.
\newblock In {\em 2016 IEEE International Conference on Image Processing (ICIP)}. IEEE, Sept. 2016.

\bibitem{ocSORT}
J.~Cao, J.~Pang, X.~Weng, R.~Khirodkar, and K.~Kitani.
\newblock Observation-centric sort: Rethinking sort for robust multi-object tracking, 2023.

\bibitem{sportsMOT}
Y.~Cui, C.~Zeng, X.~Zhao, Y.~Yang, G.~Wu, and L.~Wang.
\newblock Sportsmot: A large multi-object tracking dataset in multiple sports scenes, 2023.

\bibitem{strongSORT}
Y.~Du, Z.~Zhao, Y.~Song, Y.~Zhao, F.~Su, T.~Gong, and H.~Meng.
\newblock Strongsort: Make deepsort great again.
\newblock {\em IEEE Transactions on Multimedia}, 2023.

\bibitem{qdtrack}
T.~Fischer, T.~E. Huang, J.~Pang, L.~Qiu, H.~Chen, T.~Darrell, and F.~Yu.
\newblock Qdtrack: Quasi-dense similarity learning for appearance-only multiple object tracking, 2023.

\bibitem{memotr}
R.~Gao and L.~Wang.
\newblock Memotr: Long-term memory-augmented transformer for multi-object tracking, 2024.

\bibitem{soccerNetPaper}
S.~Giancola, M.~Amine, T.~Dghaily, and B.~Ghanem.
\newblock Soccernet: A scalable dataset for action spotting in soccer videos.
\newblock In {\em 2018 IEEE/CVF Conference on Computer Vision and Pattern Recognition Workshops (CVPRW)}, pages 1792--179210, 2018.

\bibitem{deepEIoU}
H.-W. Huang, C.-Y. Yang, J.-N. Hwang, and C.-I. Huang.
\newblock Iterative scale-up expansioniou and deep features association for multi-object tracking in sports.
\newblock {\em arXiv preprint arXiv:2306.13074}, 2023.

\bibitem{kalmanFilter}
R.~E. Kalman.
\newblock A new approach to linear filtering and prediction problems.
\newblock {\em Transactions of the ASME--Journal of Basic Engineering}, 82(Series D):35--45, 1960.

\bibitem{deepocsort}
G.~Maggiolino, A.~Ahmad, J.~Cao, and K.~Kitani.
\newblock Deep oc-sort: Multi-pedestrian tracking by adaptive re-identification, 2023.

\bibitem{SnTrackingWinner}
A.~Maglo, A.~Orcesi, J.~Denize, and Q.~C. Pham.
\newblock Individual locating of soccer players from a single moving view.
\newblock {\em Sensors}, 23(18), 2023.

\bibitem{mot16}
A.~Milan, L.~Leal-Taixe, I.~Reid, S.~Roth, and K.~Schindler.
\newblock Mot16: A benchmark for multi-object tracking, 2016.

\bibitem{motiontrack}
Z.~Qin, S.~Zhou, L.~Wang, J.~Duan, G.~Hua, and W.~Tang.
\newblock Motiontrack: Learning robust short-term and long-term motions for multi-object tracking, 2023.

\bibitem{SoccerNet2023}
SoccerNet.
\newblock Soccernet tracking.
\newblock \url{https://www.soccer-net.org/tasks/tracking}, 2023.
\newblock Accessed: 2024-06-06.

\bibitem{transtrack}
P.~Sun, Y.~Jiang, R.~Zhang, E.~Xie, J.~Cao, X.~Hu, T.~Kong, Z.~Yuan, C.~Wang, and P.~Luo.
\newblock Transtrack: Multiple-object tracking with transformer.
\newblock {\em CoRR}, abs/2012.15460, 2020.

\bibitem{deepSORT}
N.~Wojke, A.~Bewley, and D.~Paulus.
\newblock Simple online and realtime tracking with a deep association metric, 2017.

\bibitem{CBIOU}
F.~Yang, S.~Odashima, S.~Masui, and S.~Jiang.
\newblock Hard to track objects with irregular motions and similar appearances? make it easier by buffering the matching space, 2023.

\bibitem{byteTrack}
Y.~Zhang, P.~Sun, Y.~Jiang, D.~Yu, F.~Weng, Z.~Yuan, P.~Luo, W.~Liu, and X.~Wang.
\newblock Bytetrack: Multi-object tracking by associating every detection box.
\newblock 2022.

\bibitem{fairmot}
Y.~Zhang, C.~Wang, X.~Wang, W.~Zeng, and W.~Liu.
\newblock Fairmot: On the fairness of detection and re-identification in multiple object tracking.
\newblock {\em International Journal of Computer Vision}, 129(11):3069–3087, Sept. 2021.

\bibitem{centertrack}
X.~Zhou, V.~Koltun, and P.~Krähenbühl.
\newblock Tracking objects as points, 2020.

\end{thebibliography}
}

\newpage
\appendixtitleon
\appendixtitletocon
\begin{appendices}

\end{appendices}

\end{document}